\pdfoutput=1
\documentclass[11pt]{article}

\usepackage{acl}

\usepackage{times}
\usepackage{latexsym}

\usepackage{times}
\usepackage{latexsym}
\usepackage{url}
\usepackage{amsfonts}
\usepackage{multirow}
\usepackage{amssymb}
\usepackage{subfigure}
\usepackage{graphicx}
\usepackage{arydshln}
\usepackage{amsmath}
\usepackage{caption}
\usepackage{CJKutf8}
\usepackage{caption}
\usepackage{capt-of}
\usepackage{amsmath}
\usepackage{CJKutf8}
\usepackage{amssymb}% http://ctan.org/pkg/amssymb
\usepackage{pifont}% http://ctan.org/pkg/pifont
\usepackage{booktabs}
\usepackage{bbding}

\usepackage[T1]{fontenc}
\usepackage[utf8]{inputenc}

\usepackage{microtype}

\title{FactMix: Using a Few Labeled In-domain Examples to Generalize to Cross-domain Named Entity Recognition}

\author{
  Linyi Yang$^{1,4}$\thanks{\ \ Equal contribution. Random order of the authorship.}, Lifan Yuan$^{2*}$\thanks{\ \ Work done at Westlake University as an intern.}, Leyang Cui$^{3}$, Wenyang Gao$^{1}$, Yue Zhang$^{1,4}$\\
  $^{1}$ School of Engineering, Westlake University\\
  $^{2}$ Huazhong University of Science and Technology\\
  $^{3}$ Tencent AI Lab \\
  $^{4}$ Institute of Advanced Technology, Westlake Institute for Advanced Study\\
  $^{1}$\texttt{\{yanglinyi, gaowenyang, zhangyue\}@westlake.edu.cn} \\
  $^{2}$\texttt{lievanyuan173@gmail.com} \ \ $^{3}$\texttt{leyangcui@tencent.com}
 }

\begin{document}
\maketitle
\begin{abstract}

Few-shot Named Entity Recognition (NER) is imperative for entity tagging in limited resource domains and thus received proper attention in recent years. Existing approaches for few-shot NER are evaluated mainly under in-domain settings. In contrast, little is known about how these inherently faithful models perform in cross-domain NER using a few labeled in-domain examples. This paper proposes a two-step rationale-centric data augmentation method to improve the model's generalization ability. Results on several datasets show that our model-agnostic method significantly improves the performance of cross-domain NER tasks compared to previous state-of-the-art methods, including the data augmentation and prompt-tuning methods. Our codes are available at \url{https://github.com/lifan-yuan/FactMix}.

\end{abstract}

\section{Introduction}
Named Entity Recognition (NER) is a subtask of natural language processing, which detects the mentions of named entities in input text, such as location, organization, and person \cite{sang2003introduction,jointNER,cui-etal-2021-template}. It has attracted research from academia and industry due to its broadened usage in customer services and document parsing as a core task in natural language understanding \cite{nadeau2007survey, lstm-cnn-crf,lan,luke}. However, training data for NER is available only for limited domains. It has been shown that such labeled data introduces challenges for a model to generalize to new domains \cite{snell2017prototypical,bert-ner,lin2021rockner}. 

To address this problem, a line of research considers how to allow a model to effectively learn from a few labeled examples in a new target domain \cite{zhang2021flexmatch,ma2021template,das2021container,chen2022few,wangusb,wang2022freematch}. However, such methods still require manual labeling for target domains, which makes them difficult to generalize to zero-shot diverse domain settings. A different line of research in NER considers data augmentation, using automatically constructed labeled examples to enrich training data. \citet{zeng-etal-2020-counterfactual} consider using entity replacement to generate intervened new instances. We follow this line of work and consider a new setting -- how to generate NER instances for data augmentations effectively -- so that a few labeled examples in a source domain can generalize to arbitrary target domains.

\begin{figure}[t]
    \centering
    \includegraphics[trim=70 160 280 60, clip, width=.8\hsize]{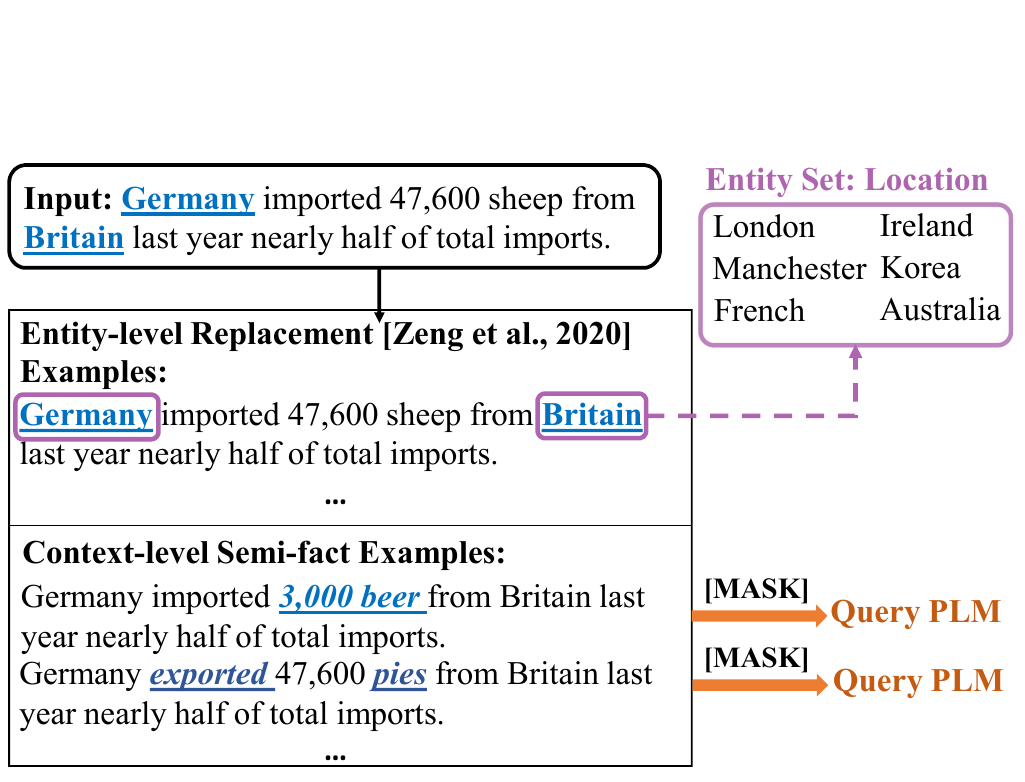}
    \caption{The demonstration of two components of FactMix, namely context-level semi-fact and entity-level semi-fact examples.}
    \label{fig:exp}
\end{figure}

% Unique challenges
Cross-domain NER poses unique challenges in practice. First, as a structured learning problem, it is essential to understand dependencies within the labels instead of classifying each token independently \cite{dai2020analysis}. While examples from different domains usually have different dependency patterns, which inevitably brings challenges for fine-tuning few-shot NER models to cross-domain tasks \cite{liu2021crossner}. Second, non-entity tokens in NER do not hold unified semantic meanings, but they could become noisy when combined with entity tokens in the training set. Such compositional generalization challenges have proven to be manifest in performance decay problems in various NLP tasks, such as sentiment analysis \cite{kaushik2019learning} and machine translation \cite{li2021compositional}, especially when faced with out-of-domain data. 

As a consequence of the challenges above, spurious patterns between non-entity tokens and labels learned by models could obstruct the generalization of few-shot NER models in cross-domain settings. For example, given that ``Jane monitored the patient's heart rate", `Jane' is labeled as a person. The NER model will learn the relationship between the word `Jane' and `monitor' for the prediction. Suppose a NER model is trained on a medical domain and tested on the movie review. The correlation between `Jane' and `monitor' could become the `spurious pattern' \cite{kaushik2019learning,yang2021exploring}. From a causal perspective, spurious correlations are caused by confounding factors rather than a cause-effect relation. 
%We leverage the masked language models to introduce out-of-context information when generating augmented examples.  

To deal with these challenges and avoid spurious patterns, we present a novel model-agnostic, two-step, rationale-enhanced approach called \textbf{FactMix}, where we care about the efficacy of data augmentations for improving in-domain and out-of-domain (OOD) performance. We aim to leverage the contrast among -- original, context-level semi-fact, and entity-level semi-fact instances -- for teaching the model to capture more causal label dependencies between entities and the context. As Figure \ref{fig:exp} shows, FactMix consists of two parts, namely context-level semi-fact generations and entity-level semi-fact instances generations. It is motivated by the natural intuition that models are much easier to learn from two-step contrastive examples compared to the one-step semi-fact augmentation \cite{zeng-etal-2020-counterfactual} \footnote{ \citet{zeng-etal-2020-counterfactual} use ``counterfactual'' to denote the setting, where augmented data contains different entities with \emph{the same} type compared with the original data. However, strictly speaking, ``counterfactual'' refers to augmented data that contains \emph{different} types of entities with a minimum change of the input that can flip the predicted label. Hence, we use semi-fact instead in our paper}.

The semi-factual generation component aims to alleviate the pitfall of non-entity tokens, which the previous data augmentation approach has not considered. We conduct synonym substitutions for non-entity tokens only. In particular, we mask the non-entity tokens, leverage the masked language models to predict the masked tokens, and replace the original tokens with predicted tokens. This replacement operation potentially introduces out-of-context information produced by the pre-trained masked language model when generating augmented examples. The entity-level semi-fact examples are generated by replacing the existing entity words in the training set. Finally, the augmented data generated by two steps will be mixed up together for training models. FactMix is a fully automatic method that does not require any additional hand-labeled data or human interventions and can be plugged for any few-shot NER models with different tuning strategies, including the standard fine-tuning and recent prompt-tuning. 

Our method supoorts a new cross-domain NER setting, which is difficult from existing work. In particular, existing few-shot NER work considers in-domain fine-tuning \cite{bert-ner} and in-domain prompt-tuning \cite{cui-etal-2021-template}. While our method also considers using only a source domain dataset for training models that generalize to target domains. Experimental results show that FactMix can achieve an average 3.16\% performance gain in the in-domain fine-tuning setting compared to the state-of-the-art entity-level semi-fact generation approach \cite{zeng-etal-2020-counterfactual} and an average 6.85\% improvement for prompt-tuning compared to EntLM \cite{ma2021template}. Improvements in such a scale hint that FactMix builds a novel benchmark. To the best of our knowledge, we are the first to explore the cross-domain few-shot NER setting using fine-tuning and prompt-tuning methods. 

%We will release our code and dataset on Github upon acceptance.

%For example, in the instance ``Jane monitored the patient's heart rate", `Jane' is labeled as a person. Suppose the training set has a large number of such patterns, the model will learn the relationship between the word `Jane' and `monitor' for the prediction. They check the monitor to see what is in each bag.

\section{Related Work}

\textbf{Cross-domain NER} focuses on transferring NER models across different text styles \cite{pan2013transfer,xu2018cross,liu2021crossner,chen2021data}. Current NER models cannot guarantee well-generalizing representation for out-of-domain data and result in sub-optimal performance. To address this issue, \citet{lee-etal-2018-transfer} continue fine-tuning the model trained on the source domain by using the data from the target domain. \citet{jointNER} jointly train NER models in both the source domain and target domain. \citet{jia2019cross} and \citet{jia2020multi} perform cross-domain knowledge transfer by using the language model. These methods rely on NER annotation or raw data in the target domain. In contrast, we propose a data argumentation method that only boosts cross-domain performance by using the source-domain corpus.

% \textbf{Out-of-domain Generalization} has gained more and more attention in recent years \cite{}. In particular, there is a line of work \cite{} focusing on the transfer learning among different text styles. We argue that current NER models cannot guarantee well-generalizing representation for out-of-domain data and result in sub-optimal performances. In terms of the cross-domain NER, \citet{jiachen} is the first method that can deal with a zero-shot learning setting for unsupervised domain adaption. At the same time, previous works still require labeled data from the target domain. Following \cite{jiachen}, we also evaluate our methods in zero-shot domain adaption settings.

\textbf{Few-shot NER} aims to recognize pre-defined named entities by only using a few labeled examples and is commonly used for evaluating structured prediction models in recent \cite{ravi2016optimization,snell2017prototypical,das2021container,luo2022mere}. \citet{label-agnostic} and \citet{nearest-neighbor-crf} propose distance-based methods, which copy the label of nearest neighbors. \citet{huang-etal-2021-shot} further investigates the efficacy of the self-training method on external data based on the distance-based methods. \citet{cui-etal-2021-template} and \citet{bert-ner} adopt prompt-based methods by using BART and BERT, respectively. These methods focus on designing few-shot-friendly models without any external guidance. In contrast, we augment both entity-level semi-fact and context-level semi-fact examples to boost the model performance on the new cross-domain few-sot setting. 

The area of \textbf{Few-shot Cross-domain Learning} is motivated by the ability of humans to learn object categories from a few examples at a rapid pace, which is called rationale-based learning. Inductive bias \cite{baxter2000model,zhang2020every} has been identified for a long time as a critical component. Benefits from the rapid development of large-scale pre-trained language models, few-shot learning, and out-of-distribution generalization become rapidly growing fields of NLP research \cite{brown2020language,Shen2021TowardsOG,chen2022few}. However, these two research directions have been separately explored in down-streaming tasks but rarely discussed together, except in the very recent study of sentiment analysis \cite{lu2022rationale}. To the best of our knowledge, we are the first to consider this setting for NER.

\textbf{Data Augmentation} through deformation has been known to be effective in various text classification tasks \cite{feng2021survey,li2022data}, such as sentiment analysis \cite{yang2021exploring,lu2022rationale} and natural language inference \cite{Kaushik2021ExplainingTE,wu2022generating}. In the task of NER, self-training has been applied to automatically increase the amount of training data \cite{wang2020adaptive}. \citet{paul2019handling} propose to combine self-training with noise handling on the self-labeled data to increase the robustness of the NER model. \citet{bansal2020self} and \citet{wang2021meta} develop self-training and meta-learning techniques for training NER models with few labels, respectively. 

%They find that self-training can be served as an effective mechanism to learn from unlabeled data, while meta-learning \cite{bansal2020self,wang2021meta} can help in mitigating errors from noisy labels. 

In addition to self-training methods, prompt-based \cite{lee2021good,ma2021template} and causal-enhanced \cite{zeng-etal-2020-counterfactual} approaches have also surfaced in this domain, which are two important baselines for our work. \citet{zeng-etal-2020-counterfactual} consider using the human intervention to generate the augmented data to improve few-shot NER models, and \citet{ma2021template,luo2022exploiting} aims to leverage the template-free prompt for boosting the performance of few-shot NER models. Nevertheless, both methods only focus on the in-domain accuracy while ignoring the cross-domain generalization of few-shot NER models.

% \textbf{Few-shot NER} is commonly used for evaluating pre-trained language models on structured predictions. Recently, it has been addressed using different approaches, including metric-learning \cite{}, meta-learning \cite{}, and data augmentation \cite{}. Metric learning usually relies on distance-based methods to find the nearest neighbors. Their assumption is that name entity patterns are similar between the source domain and the target domain, and thus help transfer knowledge. Based on the distance-based methods, \citet{huang2020} investigates the efficacy of the self-training method on external data for yielding better results. While \cite{cui2021} designs a template-based method without external data by using BART. Meta-learning aims to use the episodic classification paradigm to simulate few-shot settings for learning inductive bias on distribution of similar tasks that can be transferred in different domains \cite{}. Of particular relevance to this work is the data augmentation, which has been known to be effective in text classification tasks \cite{}. 

\section{Settings}
\begin{figure}[t]
    \centering
    \includegraphics[width=.8\hsize]{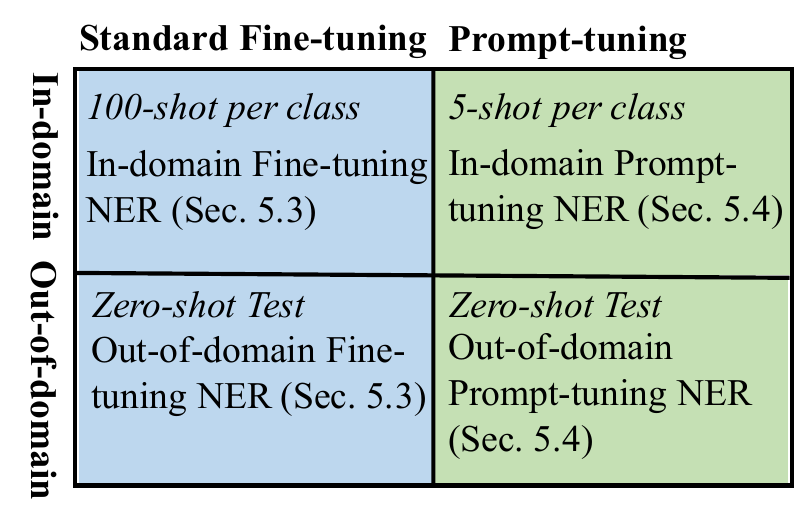}
    \caption{The categorization of experiment settings.}
    \label{fig:category}
\end{figure}

\begin{figure*}[t]
    \centering
    \includegraphics[width=.7\hsize]{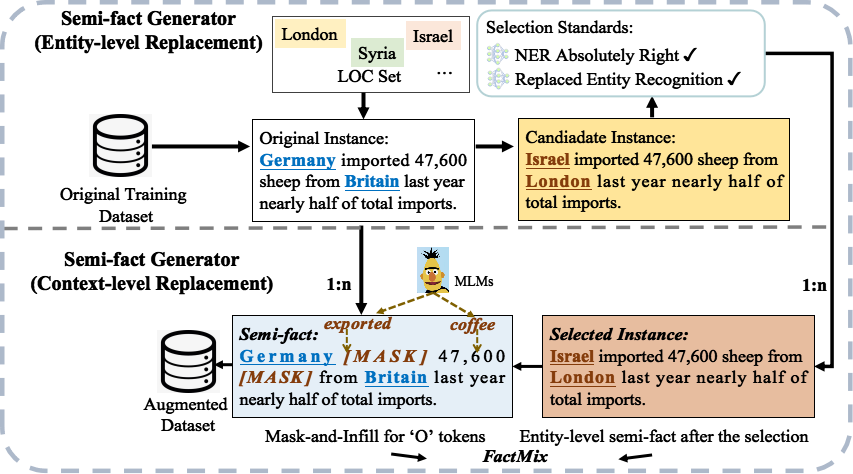}
    \caption{The pipeline of the two-step FactMix approach operated on the source domain, which consists with entity-level semi-fact generations and context-level semi-fact generations.}
    \label{fig:pipeline}
\end{figure*}

% We investigate NER under several settings as shown in Figure \ref{fig:category}, including standard fine-tuning, out-of-domain fine-tuning, in-domain prompt-tuning, and out-of-domain prompt-tuning settings. 

We investigate the effectiveness of FactMix using different methods under several settings. We first introduce task settings in Section~\ref{sec:settings}, then show the standard fine-tuning method and prompt-based method in Section~\ref{sec:ft} and Section~\ref{sec:prompt}, respectively.

\subsection{Task Settings}
\label{sec:settings}
The input of the NER system is a sentence \(\mathbf{x}=x_{1}, \ldots, x_{n}\), which is a sequence of $n$ words and the output is a sequence of NER tags \(\mathbf{y}=y_{1}, \ldots, y_{n}\), where \(y_{i} \in \mathcal{Y}\) for each word and \(\mathcal{Y}\) is selected from a pre-defined label set\(\{B-X, I-X, S-X, E-X ... O\}\). \(B, I, E, S\) represent the beginning, middle, ending, and single-word entities, respectively. \(X\) indicates the entity type, such like \(PER\) and \(LOC\), and \(O\) refers to the non-entity tokens. We use \(\mathcal{D}_{ori}\) and \(\mathcal{D}_{ood}\) to represent the original dataset and out-of-domain dataset, respectively.

Given small labelled instances of \(\mathcal{D}_{ori}\), we first train a model \(\mathcal{M}_{ori}\) through the standard fine-tuning method. We test the performance of \(\mathcal{M}_{ori}\) on \(\mathcal{D}_{ood}\) and \(\mathcal{D}_{ood}\) under \textit {In-domain Few-shot Setting} and \textit{Out-of-domain Zero-shot Setting}, respectively, which can be seen in Figure 2.

\subsection{Standard Fine-tuning Methods}
% \paragraph{Standard Fine-tuning}
\label{sec:ft}
% In order to fine-tune pre-trained language models for few-shot NER under the standard fine-tuning strategy, 

Following \citet{devlin2018bert}, we feed contextualized word embeddings into a linear classification with the softmax function to predict the probability distribution of entity types. Formally, we first feed the input \(\mathbf{x}\) into the feature encoder \(PLM_{\theta}\) to get the corresponding contextualized word embeddings \(\mathbf{h}\):
\begin{equation}
\mathbf{h}=PLM_{\theta}(\mathbf{x}),
\end{equation}
where \(\mathbf{h}\) is the sequence of contextualized word embeddings based on pre-trained language models (PLMs), i.e., BERT \cite{devlin2018bert} and RoBERTa \cite{liu2019roberta}. We optimize the cross entropy loss \(\mathcal{L}_{NER}\) by using AdamW \cite{loshchilov2018decoupled}, which is formulated as:

\begin{equation}
    \mathcal{L}_{NER}= -\sum_{c=1}^{N} y_{o, c} \log \left(p_{o, c}\right),
\end{equation}
where \(N\) is the number of classes, \(y\) is the binary indicator (0 or 1) depending on if the gold label \(c\) is the correct prediction for observation \(o\), and \(p\) is the predicted probability for observation \(o\) of \(c\).

\subsection{Prompt-tuning Methods}
\label{sec:prompt}
Prompt-tuning NER reformulates classification tasks by using the mask-and-infill technique based on human-defined templates to generate label words. We perform the template-based and template-free prompt tuning as two additional experimental scenes to verify the validity of our method. Unlike the standard fine-tuning, no new parameters are introduced in this setting.

\paragraph{Template-based Approach} Formally, we adopt the prompt template function \(F_{\text {prompt}}(\cdot)\) proposed by a very recent work \cite{ma2021template} to converts the input \(\mathbf{x}\) to a prompt input \(x_{\text {prompt }}=F_{\text {prompt}}(x)\), and pre-defined label words \(\mathcal{P}\) from the label set \(\mathcal{Y}\) are generated through a mapping function \(\mathcal{M}: \mathcal{Y} \rightarrow \mathcal{P}.\) In particular, two slots need to be infilled for each instance: the input slot [X] is filled by the original input \(\mathbf{x}\) directly, and the prompt slot [Z] is filled by the label word. To be note that [Z] is predicted by the masked language model (MLM) for prompt-based tuning in this work. The probability distribution over the label set \(\mathcal{Y}\) can be optimized by the softmax function for predicting masked tokens using pre-trained models.

\paragraph{Template-free Approach} In order to reduce the computational cost of the decoding process for template-based prompt tuning, \citet{ma2021template} propose an entity-oriented LM (EntLM) objective for fine-tuning NER. Following \citet{ma2021template}, we first construct a label word set \(\mathcal{P}_f\) by the label word engineering, which is also connected with the label set through a mapping function \(\mathcal{M}: \mathcal{Y} \rightarrow \mathcal{P}_f. \) Next, we replace entity tokens at entity positions with corresponding label word \(\mathcal{M}(y_i)\). Finally, the target input can be represented as \(\mathbf{x^{Rep}}=\left\{x_{1}, ..., \mathcal{M}\left(y_{i}\right), ..., x_{n}\right\}\). We train the language model by maximizing the probability \(P\left(\mathbf{x^{Rep}} \mid \mathbf{x}\right)\). The loss function for generating the prompt can be formulated as:
\begin{equation}
    \mathcal{L}_{RepLM}=-\sum_{i=1}^{N} \log P\left(x_{i}=x_{i}^{Rep} \mid \mathbf{x}\right),
\end{equation}
where N is the number of classes. The initial parameters of the predictive model are obtained from pre-trained language models.

\section{Method}
FactMix automatically generates semi-fact examples for both standard fine-tuning and prompt-tuning. The pipeline of our approach is shown in Figure \ref{fig:pipeline} and is made up of three components: (1) \emph{entity-level semi-fact generator}; (2) \emph{context-level semi-fact generator}; (3) \emph{augmented data selection and mixing}. Briefly, a key innovation in this work is using a mixed semi-fact generator to improve the single entity-level data augmentation approach by adding the intermediate thinking process in a human-thinking manner.
\subsection{Semi-factual Generation}
% out-of-context
We randomly remove one \textit{O} token in each sentence. Specifically, we introduce out-of-context information by randomly masking an \textit{O} word and then filling the span using the Masked Language Model (MLM), i.e., BERT \cite{devlin2018bert}. Intuitively, we can generate numerous semi-factual samples because the MLM model can fill the masked span with multiple predictions. More importantly, choosing the number and order of the selected words is a combinatorial permutation problem. However, in practice, we find that more augmented data can not always lead to a better result; and for each semi-factual sample, we only replace one word or two-word phrase in a sentence using the top one mask-and-infill prediction of MLM.

Formally, given an input of NER as \({\mathbf{x}} = {x_1}, ... ,{x_i}, ... ,{x_n}\), where \(x_{i}\) is the chosen \textit{O} word. We first mask \(x_i\) by replacing it with the [MASK] token, and thus get \({\mathbf{x}} = {x_1}, ... ,[MASK], ... ,{x_n}\). Then we fill the [MASK] token using \textit{BERT-base-cased}\footnote{\url{https://huggingface.co/bert-base-cased}} model and finally obtain a semi-factual example \({\mathbf{x_{semi}}} = {x_1}, ... ,{x_i}', ... ,{x_n}\). For instance, as seen in Fig.~\ref{fig:pipeline}, 
\textit{sheep} may first be masked and then infilled by an out-of-context word \textit{coffee}, which can be generated by PLMs.

The intervention of the selected word may inflect the entity tag of other words and introduce extra noises into the dataset. Thus, we adopt a denoising mechanism that can filter out noisy examples by leveraging the predictive model trained on the original dataset that contains prior knowledge for NER tasks. 
Different from \citet{zeng-etal-2020-counterfactual}, which filters only those samples whose replaced entities cannot be predicted correctly, we use a stricter constraint to preserve only those samples where all tokens are predicted accurately.

\subsection{Entity-level Semi-fact Generation}
% in-context
We generate entity-level semi-fact examples by interventions on the existing entity words. Specifically, for each training sample, we randomly select one of its entity words and replace it with words of the same type in a prepared $Entity\_Base$. 
For cases where data is not extremely scarce, e.g. in the fine-tuning setting in our experiments, the $Entity\_Base$ can be constructed by extracting and categorizing all entity words in the original dataset. Otherwise, e.g. in the 5-shot prompt-tuning setting in our experiments, the $Entity\_Base$ should be constructed from other available datasets.

Formally, given the input as \({\mathbf{x}} = {x_1}, \ldots ,{x_j}, \ldots ,{x_n}\), and $x_{j}$ as the chosen entity word. We assume that the label of $x_j$ is B-LOC and extract all the B-LOC entities in the $Entity\_Base$ and denote them as B-LOC Set. Next, a word in B-LOC Set is chosen to replace $x_j$ and denoted as ${x_j}'$. In this way, the generated semi-fact sample is \({\mathbf{x_{cf}}} = {x_1}, \ldots ,{x_j}', \ldots ,{x_n}\). For example, as seen in Fig.~\ref{fig:pipeline}, the B-LOC entity word \textit{German} is replaced by \textit{Israel} in B-LOC Set. All augmented samples are labeled as the same tag with original ones for saving manual efforts. 

\subsection{Mix Up}

In the last step, we combine two types of automatically generated data by a mix-up strategy. Although the FactMix method can generate an unlimited amount of data theoretically, past experience \cite{lu2022rationale} suggests that more fact-based data instances can not always bring performance benefits accordingly.

Following \citet{zeng-etal-2020-counterfactual}, we set the maximum augmentation ratio as 1:8 for the entity-level semi-fact data generation. While for context-level semi-fact generations, we set the ratio as 1:5. The optimal augmentation ratios for these two kinds of augmentations are jointly selected by the grid search on the development set of in-domain data. Finally, we obtain the final FactMix augmented training data, which can be represented as \({\mathbf{x_{mix}}} = Concat\{{\mathbf{x_{semi}}}, {\mathbf{x_{cf}}}\}\).

% semi generation
% cf, refer to xxx
% mix

% sample trainset 100 shots
% MMI selection (ablation ?)

\section{Experiments}
\begin{table}[t]
\centering
\footnotesize
\begin{tabular}{c|ccc|c}
\hline
\multirow{2}{*}{Domain} & \multicolumn{3}{c|}{\# Instances} & \multicolumn{1}{l}{\multirow{2}{*}{Entity Types}}                                                            \\ \cline{2-4}
                        & Train      & Dev       & Test     & \multicolumn{1}{l}{}                                                                                         \\ \hline
Reuters                 & 14,987      & 3,466      & 3,684     & \multirow{6}{*}{\begin{tabular}[c]{@{}c@{}}Person,\\ Location,\\ Organization,\\Miscellaneous\end{tabular}} \\
TechNews                      & -          & -         & 2000      &                                                                                                              \\
AI                      & -          & -         & 431      &                                                                                                              \\
Literature              & -          & -         & 416      &                                                                                                              \\
Music                   & -          & -         & 456      &                                                                                                              \\
Politics                & -          & -         & 651      &                                                                                                              \\
Science                 & -          & -         & 543      &                                                                                                              \\ \hline
\end{tabular}
\caption{Statistics of datasets used in experiments.}
\label{tab:dataset_details}
\end{table}
As shown in Table 2, we conduct experiments under the scenarios of both fine-tuning and prompt-tuning, using in-domain and out-of-domain evaluations. We are also interested in better understanding the contributions of the two-step data augmentation approach when it comes to prediction performance. Thus, we consider several ablation studies to better the relative contributions of  entity-level and context-level semi-fact augmented data. Micro F1 is used as evaluation metric for all settings.

\subsection{Methodology}

\paragraph{Fine-tuning.} Given that FactMix is a model-agnostic data augmentation approach, we adopt the standard fine-tuning method based on two pre-trained models with different parameter sizes: BERT-base, BERT-large, RoBERT-base, and RoBERT-large. All backbone models are implemented on the transformer package provided by Huggingface \footnote{\url{https://huggingface.co/models}}. To fine-tune NER models in a few-shot setting, we randomly sample 100 instances per label from the original dataset to ensure that the model converges. We report the average performance of models trained by five-times training. 

\paragraph{Prompt-tuning.} We adopt the recent EntLM model proposed by \citet{ma2021template} as the benchmark for prompt-tuning. Following \citet{ma2021template}, we conduct the prompt-based experiments using the 5-shot training strategy. Again, we conduct a comparison between the state-of-the-art prompt-tuning method and several variants of FactMix. We also analyze the separate contribution of the counterfactual generator and semi-fact generator by providing an ablation study based on the the base and large versions of the BERT-cased backbone. For the standard hold-out test, we report results on both development and test sets. We also select two representative datasets for the out-of-domain test in terms of the highest (TechNews) and lowest (Science) word overlap with the original training domain (Reuters).

\begin{table}[t]
\centering
\small
\resizebox{\linewidth}{!}{
\begin{tabular}{c|c|cccc}
\hline
\multirow{2}{*}{\textbf{\begin{tabular}[c]{@{}c@{}}Dataset\end{tabular}}} & \multirow{2}{*}{\textbf{Backbone}} & \multicolumn{4}{c}{\textbf{In-domain Fine-tuning Results}}        \\ \cline{3-6} 
                                                                                     &                                    & \textbf{Ori} & \textbf{E-Semi} & \textbf{C-Semi} & \textbf{FactMix} \\ \hline
\multirow{4}{*}{\begin{tabular}[c]{@{}c@{}}CoNLL2003\\ (Dev)\end{tabular}}   & BERT-base-cased                   & 57.98        & 79.78       & 81.48             & \textbf{83.13*}   \\
                                                                                     & BERT-large-cased                 & 69.18        & 83.27       & 85.87             & \textbf{85.73*}   \\
                                                                                     & RoBERTa-base                       & 52.44        & 85.81       & 87.99             & \textbf{88.51*}   \\
                                                                                     & RoBERTa-large                      & 68.81        & 88.25       & 89.39             & \textbf{89.95*}   \\ \hline
\multirow{4}{*}{\begin{tabular}[c]{@{}c@{}}CoNLL2003\\ (Test)\end{tabular}}  & BERT-base-cased                    & 54.03        & 77.71       & 78.70             & \textbf{80.10*}   \\
                                                                                     & BERT-large-cased                   & 65.38        & 81.11       & \textbf{83.04}    & 82.65            \\
                                                                                     & RoBERTa-base                       & 48.53        & 82.74       & 85.05             & \textbf{85.33*}   \\
                                                                                     & RoBERTa-large                      & 65.70        & 85.20       & 86.84             & \textbf{86.91*}   \\ \hline
\end{tabular}}
\caption{The Micro F1 score of different models by using FactMix and related data augmentation methods -- E-Semi: Entity-level Semi-fact Generation \cite{zeng-etal-2020-counterfactual}; C-Semi: Context-level Semi-fact Generation (Ours); FactMix (Ours) -- using the in-domain few-shot fine-tuning. \(*\) indicates the statistically significant under T-test, p\(<\)0.05.}
\end{table}

\begin{table*}[t]
\centering
\small
\resizebox{\textwidth}{!}{
\begin{tabular}{c|c|cccc|c|cccc}
\hline
 \multirow{2}{*}{\textbf{Dataset}} &
 \multirow{2}{*}{\textbf{Backbone}} &  \multicolumn{4}{c|}{\textbf{Fine-tuning OOD Results}}       & \multirow{2}{*}{\textbf{Dataset}} & \multicolumn{4}{c}{\textbf{Fine-tuning OOD Results}}        \\ \cline{3-6} \cline{8-11} 
                                                                                     &                                    & \textbf{Ori} & \textbf{E-Semi}     & \textbf{C-Semi} & \textbf{FactMix} &                            & \textbf{Ori} & \textbf{E-Semi}     & \textbf{C-Semi} & \textbf{FactMix} \\ \hline
 \multirow{4}{*}{TechNews} &  BERT-base-cased  & 41.46  & 61.20          & \textbf{65.20*}   & 65.09*  & \multirow{4}{*}{Music}     & 10.46         & 19.33          & 17.59            & \textbf{19.49}  \\
                                                                        &       BERT-large-cased      & 52.63       & 67.51          & \textbf{69.98*}   & 69.28  &                            & 12.00         & 19.64          & 19.32            & \textbf{19.97*}  \\
                                                                        &        RoBERTa-base               & 44.88       & 71.83          & 73.15            & \textbf{73.62*}  &                            & 11.78         & 22.24          & 21.37            & \textbf{23.75*}  \\
                                                                     &         RoBERTa-large & 51.76       & 73.11          & \textbf{74.89*}   & 74.62  &                            & 14.44         & 21.13          & \textbf{22.93*}   & 20.96           \\ \hline             
                                     \multirow{4}{*}{AI} & BERT-base-cased                & 15.88       & 22.49          & 23.66            & \textbf{24.67*}  & \multirow{4}{*}{Politic}  & 21.38       & 41.84          & 40.82            & \textbf{43.60*}  \\
                                                                     &      BERT-large-cased             & 18.62       & 26.00          & 26.03   & \textbf{26.25*}  &                            & 29.77       & 43.37          & 42.57            & \textbf{43.84*}  \\
                                                                    &     RoBERTa-base                  & 18.63       & 32.03 & 29.79            & \textbf{32.09}  &                            & 26.81       & 44.12          & 44.09            & \textbf{44.66*}  \\
                                                                    &       RoBERTa-large               & 23.27       & 28.76*          & 29.77*            & \textbf{30.06*}  &                            & 28.56       & \textbf{45.87} & 44.36            & 45.05           \\ \hline
                                \multirow{4}{*}{Literature} & BERT-base-cased & 12.85  & 22.89          & 23.05  & \textbf{25.70*}  & \multirow{4}{*}{Science}   & 12.41       & 25.67          & 28.26            & \textbf{29.72*}  \\
                                  &     BERT-large-cased         & 17.53       & 24.96          & \textbf{26.25*}            & 25.39  &                            & 16.05       & \textbf{28.75} & 27.02            & 27.88           \\
                                                                    &       RoBERTa-base               & 15.05       & 28.21          & 27.90            & \textbf{28.89*}  &                            & 14.17       & 33.33          & 31.06            & \textbf{34.13*}  \\
                                                                    &       RoBERTa-large                & 19.20       & 25.43          & \textbf{26.76*}    & 26.30*  &                            & 17.25       & 31.36          & 29.89            & \textbf{32.39*}  \\ \hline
\end{tabular}
}

\caption{The average five times running results of Fine-tuning OOD over six datasets using various data augmentation approaches compared to the original training method (Standard Fine-tuning). E-Semi: Entity-level Semi-fact Generation \cite{zeng-etal-2020-counterfactual}; C-Semi: Context-level Semi-fact Generation (Ours); FactMix (Ours). \(*\) indicates the statistically significant under T-test, p\(<\)0.05, when compared to E-Semi.}
\end{table*}

% \begin{table}[ht]
% \centering
% \small
% \resizebox{\linewidth}{!}{
% \begin{tabular}{c|c|cccc}
% \hline
% \multirow{2}{*}{\textbf{Dataset}} & \multirow{2}{*}{\textbf{Backbone}} & \multicolumn{4}{c}{\textbf{In-domain Prompt-tuning}} \\ \cline{3-6} 
%  &  & \textbf{EntLM} & \textbf{CF} & \textbf{Semi} & \textbf{FactMix} \\ \hline
% \multirow{4}{*}{\begin{tabular}[c]{@{}c@{}}CoNLL2003\\ (Dev)\end{tabular}} & BERT-base-cased & 51.73 & 48.14 & 59.30 & \textbf{62.40} \\
%  & BERT-large-cased & 60.95 & 58.42 & 49.53 & \textbf{61.64} \\
%  & RoBERTa-base & 54.57 & 53.66 & 60.62 & \textbf{62.49} \\ 
%  & RoBERTa-large & 57.90 & 57.67 & 63.48 & \textbf{64.01} \\\hline
%  \multirow{4}{*}{\begin{tabular}[c]{@{}c@{}}CoNLL2003\\ (Test)\end{tabular}} & BERT-base-cased & 54.00 & 55.61 & 57.23 & \textbf{59.19} \\
%  & BERT-large-cased & 60.80 &  &  &  \\
%  & RoBERTa-base &  &  &  &  \\ 
%  & RoBERTa-large & & & & \\\hline
% \end{tabular}}
% \caption{In-domain Prompt-tuning Results.}
% \end{table}

\begin{table*}[ht]
\centering
\small
\resizebox{\linewidth}{!}{
\begin{tabular}{c|c|cccc|c|cccc}
\hline
 \multirow{2}{*}{\textbf{Dataset}} &
 \multirow{2}{*}{\textbf{Backbone}} &
 \multicolumn{4}{c|}{\textbf{Prompt-tuning In-domain Results}} & \multirow{2}{*}{\textbf{Dataset}} & \multicolumn{4}{c}{\textbf{Prompt-tuning OOD Results}} \\ \cline{3-6} \cline{8-11} 
 &  & \textbf{EntLM} & \textbf{E-Semi}     & \textbf{C-Semi} & \textbf{FactMix} &  & \textbf{EntLM} & \textbf{E-Semi}     & \textbf{C-Semi} & \textbf{FactMix} \\ \hline
\multirow{2}{*}{\begin{tabular}[c]{@{}c@{}}CoNLL2003\\ (Dev)\end{tabular}} & BERT-base-cased & 51.73 & 48.14 & 59.30 & \textbf{62.40*} & \multirow{2}{*}{TechNews} & 47.16 & 52.36 & 50.96 & \textbf{52.44*} \\
 & BERT-large-cased & 60.95 & 58.42 & 49.53 & \textbf{61.64*} &  & \textbf{52.53} & 48.32 & 32.48 & 48.64 \\\hline
\multirow{2}{*}{\begin{tabular}[c]{@{}c@{}}CoNLL2003\\ (Test)\end{tabular}} & BERT-base-cased & 54.00 & 55.61 & 57.23 & \textbf{59.19*} & \multirow{2}{*}{Science} & 15.70 & 18.32 & 17.28 & \textbf{18.62*} \\
 & BERT-large-cased & 60.37 & 56.49 & 58.37 & \textbf{60.80*} & & 15.32 & 15.34 & 13.01 & \textbf{16.80*}  \\\hline
\end{tabular}}
\caption{The comparison among our methods, entity-level semi-fact data augmentation, and EntLM \cite{ma2021template} using prompt-tuning hold-out test and OOD test. \(*\) indicates the statistically significant under T-test, p\(<\)0.05, when compared to EntLM.}
\end{table*}

\subsection{Datasets}
The statistics of both source domain and out-of-domain datasets are introduced in Table~\ref{tab:dataset_details}. As a common understanding, it is easy to collect a large unlabeled corpus for one domain, while the corpus size could be small for low-resource domains. Then, we introduce datasets used in experiments for in-domain tests and out-of-domain tests, respectively, as follows.

%We can observe that the training and testing size of the source domain is much larger than other domains.

\paragraph{In-domain Dataset.} We conduct the in-domain experiments on the widely used CoNLL2003 \cite{sang2003introduction} dataset with a text style of Reuters News and categories of person, location, organization, and others.   

\paragraph{Out-of-domain Datasets.} We adopt the cross-domain dataset collected by \citet{liu2021crossner} with new domains of AI, Literature, Music, Politics, and Science. Vocabularies for each domain are created by considering the top 5K most frequent words (excluding stopwords). \citet{liu2021crossner} report that vocabulary overlaps between domains are generally small, which further illustrates that the overlaps between domains are comparably small and out-of-domain datasets are diverse. Notably, since the model trained on CoNLL2003 can only predict person, location, organization, and various entities, we set all the unseen labels in OOD datasets to \textit{O}.

\subsection{Results on Few-shot Fine-tuning}
In-domain experimental results on a widely used CoNLL2003 dataset show that FactMix achieves an average \textbf{3.16\%} performance gain in the in-domain fine-tuning setting (100 instances per class) and an average \textbf{2.81\%} improvement for prompt-tuning (5 instances per class) compared to the state-of-the-art data augmentation approach. For OOD test results, FactMix increases absolute \textbf{14.19\%} F1 score in average in fine-tuning compared to \cite{zeng-etal-2020-counterfactual} and \textbf{1.45\%} increase in prompt-tuning compared to \cite{ma2021template}.

\textbf{In-domain Fine-tuning} results are presented in Table 2 under the standard fine-tuning setting, using each of the baselines (Ori) and several variations of our FactMix approach. All results average five times running with randomly training instance selections.

FactMix achieves the best performance on both development and test sets, in terms of the highest Micro F1 score, excluding that the BERT-large model can achieve the best performance using our semi-fact augmentation approach only. Furthermore, we observe that improvements introduced by variants of the data augmentation approach are relatively significant when compared to models trained without data augmentations (\textbf{25.3\%} absolute F1 improvements on average). FactMix also shows its superior performance compared to the previous state-of-the-art data augmentation method \cite{zeng-etal-2020-counterfactual} with a 2.1\% absolute improvement in average. Finally, FactMix establishes a new state-of-the-art for the data augmentation approach in the cross-domain few-shot NER. 

\textbf{Out-of-domain Fine-tuning.}
We consider the performance of few-shot NER in the context of a more challenging cross-domain setting. The micro-f1 score of pre-trained models based on different augmentation methods is shown in Table 3. We find that the performance decay in technews is relatively lower than other domains since the technews domain also holds a relatively higher overlap with the training set (Reuters News). Again, our semi-factual generation and FactMix achieve the best performance in most settings. For instance, the RoBERTa-large model trained with Semi-fact Only and FactMix can achieve 74.89\% and 74.62\% F1, respectively, compared to only 51.76\% F1 using the original training set. We also notice that all pre-trained methods manifest a significant drop in accuracy on other datasets, which share fewer overlaps with the training data than technews. For example, the RoBERTa-base model gets an \textbf{11.78\%} F1 by using the standard fine-tuning, while it can be improved to \textbf{23.75\%} with FactMix. Moreover, we can see that our methods, including Semi-fact and FactMix, achieve a significantly consistent improvement over different datasets compared to standard fine-tuning and the previous state-of-the-art method \cite{zeng-etal-2020-counterfactual}, no matter the dataset distribution gap between domains. Finally, the ablation study shows that the mix-up strategy can effectively improve the performance of fine-tuning methods in most scenarios, compared to the single semi-fact augmentation method.

\subsection{Results on Few-shot Prompt-tuning}

To further understand the benefits of FactMix, in what follows, we also consider several ablation studies based on the few-shot prompt-tuning setting (5 instances per class).

\textbf{In-domain Prompt-tuning.}
The results are shown in Table 4. We can see that FactMix achieves the best performance in 5-shot prompt-tuning on the development set and test set of CoNLL2003, compared to EntLM \cite{ma2021template} and the ablation part of FactMix. The overall Micro F1 score of prompt-tuning with FactMix is relatively lower than the results of 100-shot fine-tuning, i.e., \textbf{88.51 vs. 60.80} based on the BERT-large model. It is noteworthy that our approach shows its superior for all settings, while the previous data augmentation approach \cite{zeng-etal-2020-counterfactual} hurts the performance when using the BERT-large models, i.e., the F1 score decreases from 60.37 to 56.49 as shown in the test set. The stable performance further proves that two-step fact-based augmentations can significantly benefit NER models for both fine-tuning and prompt-tuning models.

\textbf{Out-of-domain Prompt-tuning.} 
The OOD results for prompt-tuning methods are also shown in Table 4. In general, we observe that prompt-based tuning methods have considerable potential for the cross-domain few-shot NER. While cross-domain results evaluated on the high-overlap dataset (TechNews) with the training domain are significantly higher than the low-overlap dataset (Science), i.e., \textbf{52.44 vs. 18.62} based on BERT-base. Furthermore, FactMix provides the best performance based on all of the pre-trained models, compared to EntLM and its variants. In contrast, EntLM performs better than FactMix on TechNews. It hints that our method could be more useful in a low-resource setting where the overlap between the original domain and target domain is relatively low. 

\begin{figure}[t]
    \centering
    \includegraphics[width=.85\hsize]{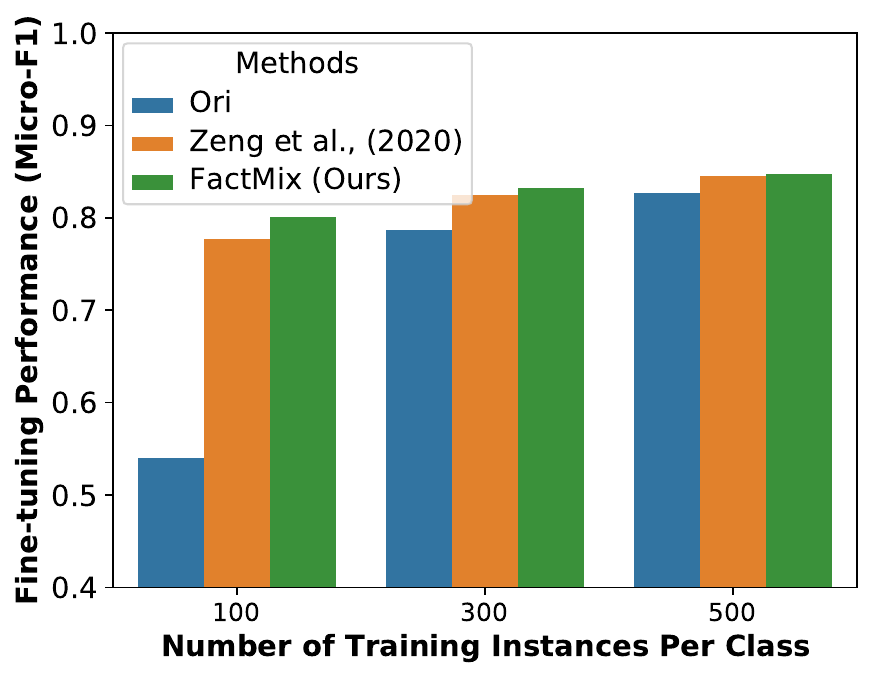}
    \caption{In-domain fine-tuning results are reported based on the BERT-base-cased model.}
    \label{fig:number}
\end{figure}

\subsection{Discussion}

Benefiting from the generalized ability of pre-trained models, FactMix achieves much improved results on the few-shot in-domain test -- 86.91\%. More importantly, it shows decent scalability when combined with fine-tuning and prompt-tuning methods. To better understand the influence of the number of initial training examples and augmentation ratios, we illustrate the comparison of in-domain fine-tuning as follows.

\textbf{The Influence of Training Samples.} 
The comparison of results based on the BERT-base-cased model is shown in Figure \ref{fig:number}. We present the results of three different methods by using the different number of training examples varying from 100 to 500. Results show that FactMix holds the best performance when the size of training examples has been set as 100, 300, and 500. We also notice that the improvements introduced by FactMix decreased as the amount of raw training data per class increased from 100 to 300 when compared to the standard fine-tuning method. Finally, our method shows its superior for all settings when compared to the previous state-of-the-art data augmentation method \cite{zeng-etal-2020-counterfactual} for Few-shot NER.

\begin{figure}[t]
    \centering
    \includegraphics[width=.9\hsize]{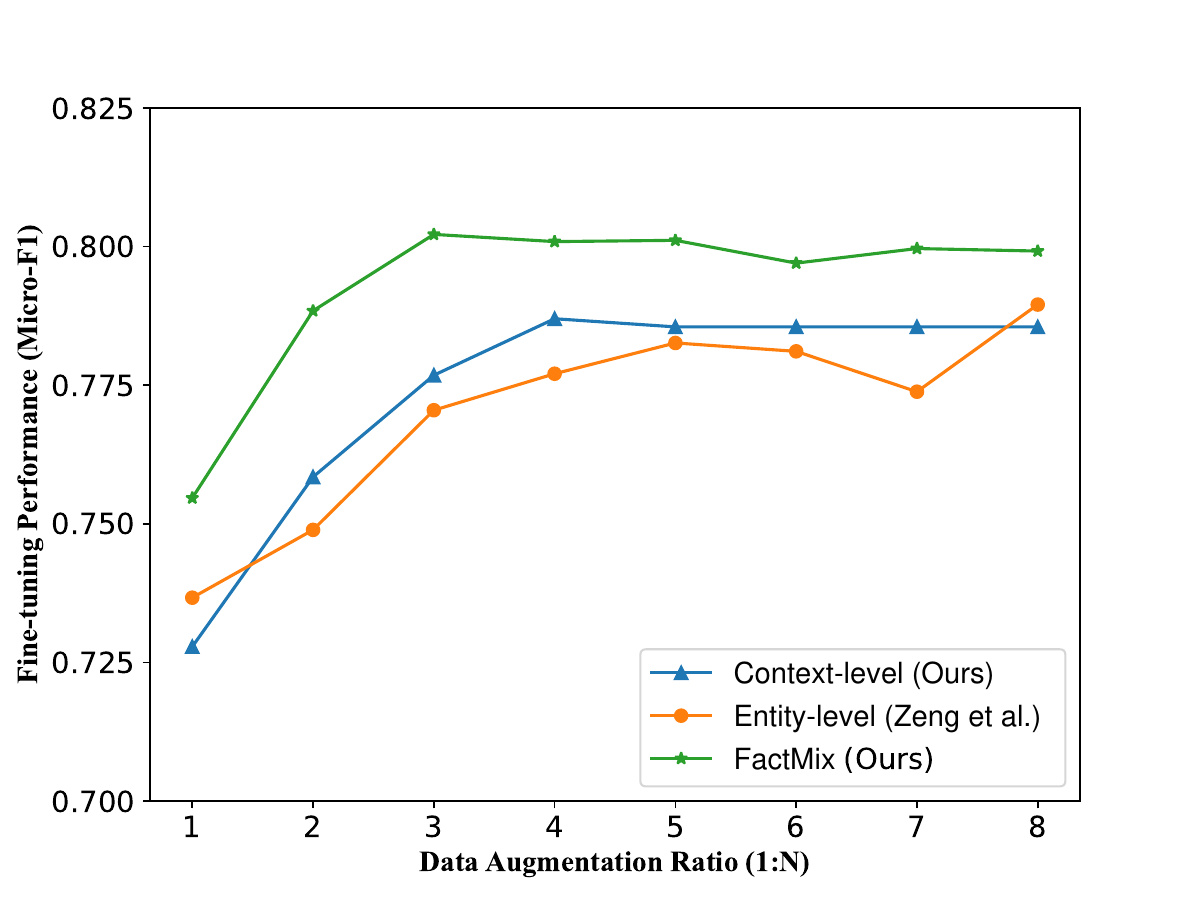}
    \caption{In-domain fine-tuning results based on BERT-base-cased using different augmentation ratios.}
    \label{fig:ratio}
\end{figure}

\textbf{The Influence of Augmentation Ratios.} In-domain fine-tuning results using different augmentation ratios are shown in Figure \ref{fig:ratio}. We consider three approaches in the evaluation, including semi-factual generation, FactMix, and the baseline method \cite{zeng-etal-2020-counterfactual}. FactMix shows its absolute performance advantage using the augmentation ratio from one to eight. In particular, Micro-F1 scores of all methods increase with the increase of the number of augmented training instances when the augmentation ratio is less than 1:4, whereas the trend of increase gradually slow down when generating examples more than 1:4.

\section{Conclusion}

We proposed a joint context-level and entity-level semi-fact generation framework, FactMix, for better cross-domain NER using few labeled in-domain examples. Experimental results show that our method can not only boost the performance of pre-trained backbones in in-distribution and OOD datasets, but also show promising results combined with template-free prompt-tuning methods. As a single data augmentation method, FactMix can be useful for different NLP tasks to enable fast generalization, i.e., relation extraction, question answering, and sentiment analysis. 

\section*{Acknowledgements}
We acknowledge with thanks the discussion with Qingkai Min from Westlake University, as well as the many others who have helped. We would also like to thank anonymous reviewers for their insightful comments and suggestions to help improve the paper. This publication has emanated from research conducted with the financial support of the Pioneer and "Leading Goose" R\&D Program of Zhejiang under Grant Number 2022SDXHDX0003. Yue Zhang is the corresponding author.

\bibliography{acl_latex}
\newpage
\appendix

\section{Appendix: Domain Distributions} \label{sec:appendix}

\begin{figure}[t]
    \centering
    \includegraphics[width=\hsize]{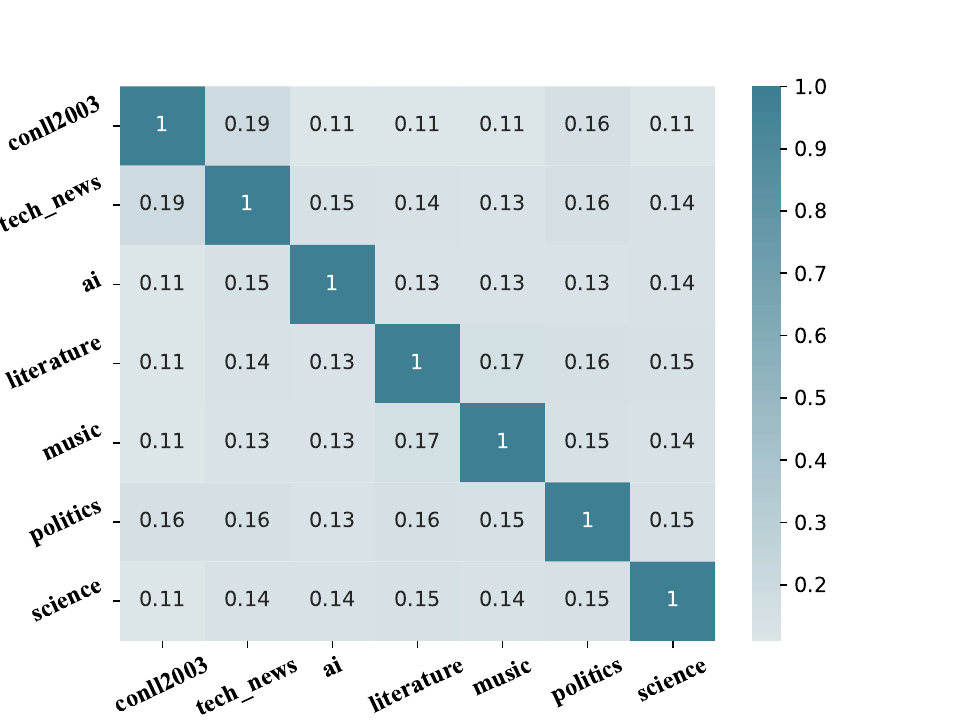}
    \caption{The word overlap between NER datasets from different domains.}
    \label{fig:overlap}
\end{figure}

The similarity between the dataset of source domain and six out-of-domain datasets is shown in Figure \ref{fig:overlap}. We find that the technical news dataset shares the highest overlap ratio with the CoNLL2003 dataset, while the science domain shares the lowest overlap. Based on that, we select TechNews and Science as two representative datasets in prompt-tuning experiments. Also, the experimental results shown in Tables 3 and 4 demonstrate that cross-domain transfer between low-overlap domains still be a challenge problem, even for FactMix.

\end{document}